\documentclass[10pt,twocolumn,letterpaper]{article}

\usepackage{cvpr}              

\definecolor{cvprblue}{rgb}{0.21,0.49,0.74}
\usepackage[pagebackref,breaklinks,colorlinks,allcolors=cvprblue]{hyperref}

\usepackage{pifont}

\newcommand{\cmark}{\ding{51}} 

\usepackage{booktabs}
\usepackage{multirow}
\usepackage{leftidx}
\usepackage{bbm}
\usepackage{floatrow}

\usepackage{algorithmic}
\usepackage{textcomp}
\usepackage{siunitx}
\usepackage{tablefootnote}
\usepackage{footnote}
\usepackage{caption}
\usepackage{makecell}
\usepackage{pifont}

\usepackage{xspace}
\usepackage{amsmath}
\usepackage{amssymb}
\usepackage{graphicx}
\usepackage{floatrow}
\usepackage{wrapfig}

\newcolumntype{P}[1]{>{\centering\arraybackslash}p{#1}}
\newcolumntype{M}[1]{>{\centering\arraybackslash}m{#1}}
\newcolumntype{L}[1]{>{\raggedright\arraybackslash}m{#1}}

\newcommand{\myparagraph}[1]{\vspace{2pt}\noindent{\bf #1}}

\def\eg{\emph{e.g.,}\xspace} 

\def\ie{\emph{i.e.,}\xspace}

\def\etal{\emph{et al.}}

\usepackage{array} 
\usepackage{caption} 
\def\paperID{13} 
\def\confName{CVPR}
\def\confYear{2025}

\title{Exploring Modality Guidance to Enhance VFM-based Feature Fusion for UDA in 3D Semantic Segmentation}

\def\authorBlock{
Johannes Spoecklberger$^{1}$\quad
Wei Lin$^{2}$\quad
Pedro Hermosilla$^{3}$\quad \\
Sivan Doveh$^{4}$\quad
Horst Possegger$^{1}$\quad
M. Jehanzeb Mirza$^{5}$\quad \\
 $^1$Institute of Visual Computing, Graz University of Technology\quad  \\
  $^2$JKU LINZ\quad
 $^3$TU Wien\quad 
 $^4$IBM Research\quad
 $^5$MIT CSAIL\quad  \\
 \tt \small j.spoecklberger@tugraz.at
}

\author{\authorBlock}

\begin{document}
\maketitle
\begin{abstract}
Vision Foundation Models (VFMs) have become a de facto choice for many downstream vision tasks, like image classification, image segmentation, and object localization.
However, they can also provide significant utility for downstream 3D tasks that can leverage the cross-modal information (\eg from paired image data).
In our work, we further explore the utility of VFMs for adapting from a labeled source to unlabeled target data for the task of LiDAR-based 3D semantic segmentation.
Our method consumes paired 2D-3D (image and point cloud) data and relies on the robust (cross-domain) features from a VFM to train a 3D backbone on a mix of labeled source and unlabeled target data.
At the heart of our method lies a fusion network that is guided by both the image and point cloud streams, with their relative contributions adjusted based on the target domain.
We extensively compare our proposed methodology with different state-of-the-art methods in several settings and achieve strong performance gains. 
For example, achieving an average improvement of 6.5 mIoU (over all tasks), when compared with the previous state-of-the-art. 

\end{abstract}      
\section{Introduction}
\label{sec:intro}
In an ideal world with abundant resources, we could potentially manually label all the distributions present in our visual world and train neural networks to perform robustly across diverse environments. 
The success of supervised training is well-known and has been extensively studied in the literature~\cite{resnet, wide-resnets, convnet}. 

However, in the real world, the concept of \textit{abundance} is usually non-existent. 
Instead, we face several constraints -- monetary, time, or human resources, among others -- that demand we employ all the available resources optimally.

Efficiency and cost reduction become even more critical in domains like autonomous driving, where vehicles ideally need to operate safely in a range of environments (\ie data distributions). 
Labeling data for each of these distributions, particularly for dense prediction tasks like 3D semantic segmentation, can quickly become prohibitively expensive due to the need for per-point or per-voxel annotations. 

\begin{figure}[t!]
    \centering
    \includegraphics[width=\linewidth]{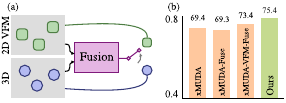}
    \caption{\textbf{(a)} Cross modal learning with frozen 2D (VFM) backbone features using a learned fusion representation. Fusion networks can lead to a suboptimal feature utilization and unwanted modality bias on the target domain. 
    Therefore, we propose regularizing the fusion by the most effective modality in a certain environment (\eg based on lighting conditions).
    \textbf{(b)} mIoU Comparison of xMUDA with different fusion variants and Ours on NuScenes: USA $\to$ Singapore.  }
    \label{fig:teasure_figure}
\end{figure}

3D Unsupervised Domain Adaptation (UDA) offers a practicable solution by focusing on adapting a neural network trained in a supervised manner on a labeled source domain to an unlabeled target domain. 
This approach leverages information from annotated 3D (source) data to address the variability across domains without requiring manually annotated labels from each new target domain. 
This paper builds on the principles of 3D domain adaptation, aiming to bridge the gap between labeled and unlabeled 3D environments, particularly addressing challenges inherent in dense 3D prediction tasks, related to fine-grained understanding of the 3D scene.

Ideally, an autonomous vehicle can employ cues from both the 2D and 3D data. 
Although different modalities, they can offer complementary cues that can enhance the perception abilities of autonomous driving systems. 
Jaritz~\etal~\cite{jaritz2022cross} proposed a seminal work 
on using the complementary information from the two modalities for the task of 3D domain adaptation for semantic segmentation. 
This method inspired a set of subsequent works~\cite{cardace2023exploiting,wu2023_bidrectional_fusion,wu2024_fusion_then_distill2}, which combine data from different domains on a similar multi-modal design.
However, these approaches usually rely on training a 2D feature extractor separately, thus, requiring disjoint training of 2D and 3D backbones, increasing the computation cost significantly as compared to uni-modal setups. 

In our work, motivated by the recent strong improvements offered by vision foundation models (VFM) on dense prediction tasks \cite{oquab2023dinov2, ranzinger2024radio}, 
we tackle the multi-modal 3D-UDA task by employing these powerful models.
However, leveraging these VFMs (\eg~\cite{ranzinger2024radio}) effectively for cross-modal learning for 3D UDA comes with the problem of optimally fusing the information from both modalities. To address this,
at the heart of our approach lies a
fusion refinement network, which fuses the information obtained from 2D and 3D modalities into a combined representation.

Although such fusion schemes are studied in~\cite{jaritz2019xmuda, li2023mseg3d}, our experiments show that training these fusion methods can lead to an over-reliance on features that perform well on the source domain but fail to adapt to the target domain.
Moreover, despite the growing popularity of multi-modal feature fusion, few works have focused on designing fusion modules that specifically address modality-specific challenges that for example arise under degrading weather and low light conditions~\cite{brodermann2024condition}.

To address this issue, we propose to \textit{adaptively} regularize the fusion network based on a simple intuition: in some environments, the imaging modality may be more robust while in others, reliance on the 3D modality may be more beneficial. 
To this end, we guide the fusion refinement network through adaptive predictive distillation from each modality, based on environmental factors (e.g., lighting conditions). This regularization effectively promotes learning to select the modality to rely on, thereby enabling more robust multi-modal domain adaptation.
An overview of our approach is also laid out in Figure~\ref{fig:teasure_figure}.
We evaluated our proposal on the four common DA tasks, showing improved performance over traditional (fusion) methods, achieving SOTA or comparable performance on all tasks without requiring the training of an additional 2D encoder.

\begin{figure*}[ht] 
    \centering
    \includegraphics[width=1\textwidth]{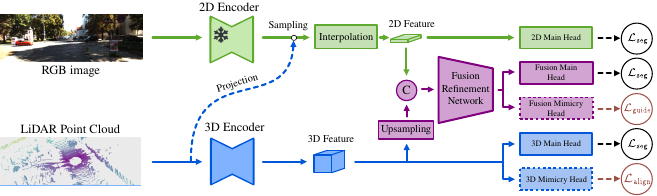} %
    \caption{Our architecture for the cross-modal learning consists of a Vision Foundation Model (VFM) as the 2D encoder and a 3D SparseConvNet as the 3D encoder. We use multiple \textit{main task heads} of semantic segmentation and \textit{mimicry task heads} for cross-modal alignment. Besides the 2D branch (green) and 3D branch (blue), we employ a fusion branch (purple) where we concatenate the 2D and 3D feature as the fusion feature.  The network is trained with the supervised loss $\mathcal{L}_\text{seg}$, and the self-supervised cross-modal learning losses $\mathcal{L}_\text{align}$ (the fusion branch guiding the 3D branch) and $\mathcal{L}_\text{guide}$ (the fusion branch guided by the 2D or 3D branch). 
    }
    \label{fig:overview}
\end{figure*}
\section{Related Work}

Our method is related to works that study UDA for 3D semantic segmentation, VFMs, and approaches that propose different fusion schemes for cross-modal learning.

\myparagraph{UDA for 3D Semantic Segmentation.}
UDA for 3D semantic segmentation has been extensively studied in recent years.

A major group of methods can be categorized as learning domain-invariant feature representations. Earlier works in 3D UDA focused on minimizing statistical feature discrepancies \cite{sun2016deep, morerio2018minimal}. Next, adversarial training is another widely used approach to learn invariant representations \cite{ganin2016domain, barrera2021cycle, Yuan2023PrototypeGuidedMA}. Another line of research utilizes self-supervision to learn more domain invariant representations by constructing a label-free auxiliary optimization goal that is often modality specific \cite{yi2021complete, michele2024saluda}. Domain mapping approaches can be seen as another high level category to handle domain shifts, in which the target domain is transferred to the source domain \cite{choi2018stargan, hoffman2018cycada}. Other approaches include model adaptation via adapting batch normalization statistics \cite{michele2024saluda, mirza2022norm, li2018adaptive}, pseudo-labeling \cite{lee2013pseudo} or self-ensembling \cite{laine2016temporal}.

In multimodal DA, techniques are characterized by performing adaptation across modalities. xMUDA \cite{jaritz2019xmuda} is a seminal work that introduces a dual classification head structure for predictive feature alignment across the 2D and 3D modalities and serves as the fundament for many works on 3D UDA cross-modal semantic segmentation. 
MM2D3D~\cite{cardace2023exploiting} extends the work by adding 3D depth to the 2D encoder, showing significant performance improvement.
DsCML~\cite{peng2021sparsetodense} proposes sparse-to-dense feature matching, aligning 3D point features with a dynamically selected 2D region. Further, they propose cross-modal adversarial learning to narrow the domain gap, which was also employed by Liu~\etal~\cite{liu2021auda}.
Xing~\etal~\cite{xing2023cross} propose a neighborhood feature aggregation network and the usage of contrastive learning for feature alignment.
Zhang~\etal~\cite{zhang2022exclusivelearning} propose modality-exclusive self-supervised learning tasks for the 2D and 3D branches to improve the exploitation of modality-specific characteristics. 
In this work, we propose a method to further enhance the 2D-3D DA capabilities enabled by VFMs. Distinct from others, we design a three-stream network architecture composed of 2D, fusion, and 3D branches, where the single-modality streams guide the fusion branch adaptively.

\myparagraph{Vision Foundation Models (VFMs).}
Vision Foundation Models, such as CLIP~\cite{radford2021learning}, DINO~\cite{caron2021emerging_dinov1}, and SAM~\cite{kirillov2023segment}, represent general-purpose frameworks trained on large-scale visual data, capable of performing a wide range of tasks, including image classification, object detection, segmentation and even cross-modal applications. 
Prior works such as Vision Transformer (ViT)~\cite{dosovitskiy2021image} laid the groundwork by demonstrating how transformer-based architectures could rival CNNs in vision tasks when trained on large datasets. 
Models like CLIP~\cite{radford2021learning} and ALIGN~\cite{jia2021scaling} exploited multi-modal capabilities showing impressive performance by learning from paired text-image data. 
DINO~\cite{caron2021emerging_dinov1} and BEiT~\cite{bao2021beit} refined the application of self-supervised learning in vision, achieving high accuracy without the need for labeled data. 
AM-RADIO~\cite{ranzinger2024radio} distills a unified efficient backbone from multiple foundation models (in a teacher-student learning framework) which outperforms its teachers across several downstream tasks.

Recently, works on DA leveraging VFMs have demonstrated strong performance for the task of 3D UDA~\cite{peng2023learning, xu2024visual, cao2024mopa}. 
Approaches to exploit VFMs can be roughly divided into works that exploit feature distillation, predictive distillation, and mask priors. 
Peng~\etal~\cite{peng2023learning} apply feature distillation via cosine similarity. 
In addition, they utilize the masking capabilities of SAM for instance-level augmentation. 
Xu~\etal~\cite{xu2024visual} employ the semantic-aware segmentation model SEEM~\cite{zou2024segment} for pseudo label refinement and mix source and target point clouds by selecting points guided by the SAM masks. Cao~\etal~\cite{cao2024mopa} exploit SAM mask priors for rare objects that are inserted in the scene tackling the long-tailed class distribution. 
Our method follows the predictive alignment paradigm, where we demonstrate that advancements in VFMs can have a direct correlation with the improvements obtained for the task of 3D UDA.

\myparagraph{Feature Fusion in 3D DA.}
xMUDA~\cite{jaritz2019xmuda} proposed a variant with late fusion of 2D and 3D features which outperforms all the unimodal baselines. 
A recent line of works \cite{wu2023_bidrectional_fusion, wu2024_fusion_then_distill2} also divert their attention toward distillation based on fused feature representations. 
Similar to our work, Wu~\etal~\cite{wu2023_bidrectional_fusion} utilizes a fusion module with a separate classifier to transfer knowledge via predictive distillation to the 2D and 3D modality. 
FtD++~\cite{wu2024_fusion_then_distill2} relies on a two-stream pipeline where the 2D stream is fused with 3D information, which is then used for cross-modal learning with the 3D stream. 
The recent UniDSeg \cite{wu2024unidseg} enhances the frozen VFM  with learnable blocks between encoder layers allowing fine-tuning and the integration of range image information.

In our approach, we treat the fusion as a distinct network trained with supervision, offering us more degrees of freedom and better adaptability to the task of interest. 
As opposed to the xMUDA fusion variant, we explicitly optimize the 3D and 2D stream via classifier heads and utilize these for guiding the distribution of the fusion network. 
Similarly, in relation to \cite{wu2023_bidrectional_fusion}, our method explicitly regularizes the fusion representation using a guiding signal to bias the fusion toward a specific modality. This regularization is a key component of our approach and is designed to enhance performance under domain shift when a modality preference is available, while still improving robustness in its absence by promoting prediction consistency on the target domain.

The recent work in \cite{brodermann2024condition} also identifies the need for non-uniform sensor fusion. In their work the fusion module is used as a central component to overcome sensing difficulties in a shared feature space. Their setup however requires scene attribute descriptions to learn a condition token which guides a windowed cross attention fusion over all modalities in the image space.

\section{Method}
We introduce our cross-modal DA approach and start with a pipeline overview in Section~\ref{subsec:overview}.  Then, we present our cross-modal fusion schema in Section~\ref{subsec:fusion}, provide details on cross-modal learning in Section~\ref{subsec:cross-modal-learning}, and conclude with details on the training objectives in Section~\ref{subsec:training-objectives}.

\subsection{Overview}
\label{subsec:overview}

\myparagraph{2D-3D Domains.} 
In the 2D-3D multimodal UDA setting, we have labeled source data and unlabeled target data from the 2D and 3D modalities. The task is to optimize the final segmentation prediction on the target data. 
We denote the source domain as $S = \{x_i^{2D}, x_i^{3D}, y_i|i\in I_S\}$ and the target domain as $T = \{x_i^{2D}, x_i^{3D}|i\in I_T\}$, where $x_i^{2D}$ and $x_i^{3D}$ denote the $i$-th image and point cloud with $y_i$ being the 3D segmentation label, while $I_S, I_T$ represent the set of sample indices of the source and target domain.

\myparagraph{Network Architecture.} An overview of our network architecture is depicted in Figure~\ref{fig:overview}. 
We tackle the task of 2D-3D cross-modal DA with two encoders to process data from the 2D and 3D modalities separately.  
Specifically, we encode the RGB images $x^{2D}$ via a pretrained and frozen 2D Vision Foundation Model (VFM) $F^{2D}(\cdot;\theta^{2D}_F)$ which yields a coarse patch feature map.
In the 3D branch, we employ a 3D SparseConvNet~\cite{graham2018sparseconvnet} denoted as $F^{3D}(\cdot;\theta^{3D}_F)$ to encode the point clouds $x^{3D}$ into 3D features.
In the 2D-3D cross-modal task, the 2D features are computed with the guidance of the input point cloud. Specifically, we first project 3D points onto the 2D image. Based on the projected pixel positions, we compute the pixel-wise 2D feature representation from patch features via bilinear interpolation.  

For the task of cross-modal semantic segmentation, we introduce several task heads. 
We first denote the \textit{main segmentation heads} for the 2D and 3D branch as $C^{2D}(\cdot; \theta^{2D}_C)$ and $C^{3D}(\cdot; \theta^{3D}_C)$ separately. Further, we follow~\cite{jaritz2019xmuda} and introduce the \textit{mimicry heads} that are additional segmentation task heads designed only for prediction alignment across different branches. The prediction alignment between dual heads across modalities are shown to be more parameter-robust than the case of a single head per modality~\cite{jaritz2022cross}.
Specifically, we introduce the mimicry head 
in the 3D branch which is denoted as $C^{3D}_\text{mmc}(\cdot; \theta^{3D}_{C_\text{mmc}})$.  Note that we do not have a mimicry head for the 2D branch as we keep the 2D VFM frozen during training.  

\subsection{Fusion Branch}\label{subsec:fusion} 
In order to enable cross-modal learning with enhanced guidance, we employ a fusion branch where we concatenate the 2D and 3D feature as the fusion feature $x^{fuse} = \text{concat}(F^{2D}(x^{2D};\theta^{2D}_F), F^{3D}(x^{3D};\theta^{3D}_F))$. Further, we employ an MLP as the fusion refinement network $F^{fuse}(\cdot;\theta^{fuse}_F)$ to refine the fusion feature $x^{fuse}$. 

Correspondingly, we construct the main segmentation head and the mimicry head for the fusion branch, denoted as $C^{fuse}(\cdot; \theta^{fuse}_C)$ and $C^{fuse}_\text{mmc}(\cdot; \theta^{fuse}_{C_\text{mmc}})$. The fusion branch is illustrated in the purple components in Figure~\ref{fig:overview}. 
During training with cross-modal alignment, our fusion branch provides the guiding signal for the 3D network. In the meanwhile, the prediction in the fusion branch is also regularized by the hypothesis in the 2D and 3D modalities. We elaborate the interaction between the fusion branch and the 2D or 3D branch in Section~\ref{subsec:cross-modal-learning}. We also ablate the impact of the fusion branch in Table~\ref{tab:fusion_alignment_symetry}. 
To obtain the final segmentation results we take the softmax average of the predictions from the 3D main head and the fusion main head.

\subsection{Cross-Modal Learning with Fusion} 
\label{subsec:cross-modal-learning}
We employ the fusion network as a cross-modal learner, trained under supervision to integrate 2D and 3D representations. To mitigate the susceptibility to domain shift, we introduce additional regularization through the 2D and 3D branches, resulting in a three-branch architecture that enables robust cross-modal learning.

Following the practice of cross-modal UDA~\cite{cardace2023exploiting, peng2023learning}, we apply the KL divergence to encourage a mimicry distribution $P_\text{mmc}$ (predictions from a mimicry head) to mimic a main head guiding distribution $P_\text{main}$ (predictions from a main head). This is shown to be more parameter-robust than the cross-modal learning with only one task head per modality~\cite{jaritz2022cross}. 
This self-supervision is applied on both source and target data, \ie
\begin{equation}
\mathcal{L}_{\text{KLD}}^{(D)}(P_\text{main},P_\text{mmc})  = \sum_{i\in I_D}  P_\text{main}(x_{i}) \log \frac{P_\text{main}(x_{i})}{P_\text{mmc}(x_{i})},
\label{eq:KL_div}
\end{equation} 
where $I_D$ is the corresponding domain sample index set.

In our empirical study, we realize that the fusion branch shows consistent improved prediction performance over the 2D branch, and therefore perform cross-modal alignment between the fusion branch and the 3D branch. We first apply the KL divergence encouraging the 3D mimicry distribution $p^{3D}_\text{mmc}$ to mimic the fusion main head distribution $p^{fuse}_\text{main}$, \ie 
\begin{equation}\label{eq:cross_align_3D_to_fuse}
\mathcal{L}_{\text{align}}^{(D)}  = \sum_{i\in I_D}  p^{fuse}_\text{main}(x_{i}) \log \frac{p^{fuse}_\text{main}(x_{i})}{p^{3D}_\text{mmc}(x_{i})}.
\end{equation} 

As the fusion branch is only supervised on the source domain, it is susceptible to domain shift from both modalities. To address this, we further regularize the fusion prediction to encourage consistency with the hypothesis of either the 2D or 3D modality on the fusion branch. \ie
\begin{multline}\label{eq:cross_align_fuse_to_2D_or_3D}
\mathcal{L}_{\text{guide}}^{(D)}  = 
\lambda\cdot\mathcal{L}_{\text{KLD}}^{(D)}(p^{2D}_\text{main},p^{fuse}_\text{mmc})~+ \\
(1-\lambda)\cdot \mathcal{L}_{\text{KLD}}^{(D)}(p^{3D}_\text{main},p^{fuse}_\text{mmc})
\end{multline}  
Here, $\lambda$ is the coefficient 
which biases the fusion towards the 2D or 3D branch and is chosen depending on the type of environmental conditions that may be expected in the task we aim to solve.   
We empirically demonstrate that this can lead to a more modality-biased fusion representation in Section~\ref{sec:experiments}.

 \subsection{Overall Training Objectives}
 \label{subsec:training-objectives}

For clarity, we denote the predicted logits from a main head as $p^m_\text{main}$ and logits from a mimicry head as $p^m_\text{mmc}$ where $p^m = C^m( F^m( x^m ; \theta^m_F)  ; \theta^m_C), m\in \{2D, 3D, fuse\}$. 
 
\myparagraph{Supervised Learning.} For each of the 2D, 3D, and fusion branches, we perform the supervised segmentation task on the three main heads, \ie
\begin{equation}\label{eq:supervised_loss}
\mathcal{L}_{\text{seg}}^{(D)}(x^m, y) =  \sum_{i\in I_D} - y_{i} \log\left( {p^m_\text{main}}_{(i)} \right),
\end{equation}
where $m\in \{2D, 3D, fuse\}$. In the following, we denote the domain as $D\in \{S, T\}$. For the supervised learning on ground truth labeled source data, we have D = S.

\myparagraph{Overall.} The overall objective is the sum of the three supervised segmentation losses on source (Eq.~\eqref{eq:supervised_loss}), and the two cross-modal losses on source and target (Eq.~\eqref{eq:cross_align_3D_to_fuse}, \eqref{eq:cross_align_fuse_to_2D_or_3D}): 

\begin{multline}\label{eq:1st_stage_overall}
 \min_\theta \; \frac{1}{|I_T|}  
  \lambda_T (\mathcal{L}_{\text{align}}^{(T)} + 
  \mathcal{L}_{\text{guide}}^{(T)} ) 
  + \frac{1}{|I_S|}  \lambda_S (\mathcal{L}_{\text{align}}^{(S)} + 
  \mathcal{L}_{\text{guide}}^{(S)} ) \\
   + \frac{1}{|I_S|}     
   \sum_{m \in M} \mathcal{L}_{\text{seg}}^{(S)}(x^m, y) 
\end{multline}
where $\theta = \{ \theta^{3D}_F, \theta^{fuse}_F, \theta^{2D}_C, \theta^{3D}_C, \theta^{fuse}_C, \theta^{3D}_{C_\text{mmc}}, \theta^{fuse}_{C_\text{mmc}}  \}$ and $M$ = {\{2D, 3D, fuse\}}.  

\myparagraph{Additional Stage with Self-Training.} To further boost the adaptation performance, we follow the practice of self-training in domain adaptation~\cite{zou2019confidence,liu2021cycle,mei2020instance,chen2020self}, and conduct the second stage by adding the supervised loss on pseudo-labeled target data. Specifically, we compute the pseudo labels by averaging the softmax scores from the fusion and the 3D branch in the first-stage model: 
\begin{multline}
\hat{y}_i = \frac{1}{2}\biggl(\text{softmax}\left(p^{fuse}_\text{main}(x_i)\right) +\\ \text{softmax}\left(p^{3D}_\text{main}(x_i)\right)\biggr), i\in I_T.
\end{multline}

In the second stage, the overall loss is the objective from stage 1 (Eq.~\eqref{eq:1st_stage_overall}) together with the additional supervised loss on target $\frac{1}{|I_T|} \lambda_{PL}   
   \sum_{m \in M} \mathcal{L}_{\text{seg}}^{(T)}(x^m, \hat{y})$.

\section{Experiments}
\label{sec:experiments}
In this section we describe our conducted experiments along with the setup following the commonly used 3D UDA evaluation based on \cite{jaritz2022cross}.
\subsection{Datasets}
We evaluate our method on the widely used nuScenes \cite{caesar2020nuscenes}, SemanticKITTI~\cite{behley2019semantickitti} (SK), VirtualKITTI~\cite{gaidon2016virtualkitti} (VK) and the A2D2 dataset~\cite{geyer2020a2d2}.  All datasets provide a synchronized and calibrated setup that allows point to pixel projection. For simplicity and comparability, only front image sensor data is utilized.
The DA tasks feature four domain shift scenarios, aiming to evaluate diverse scenarios: \emph{USA $\to$ Singapore} evaluates adaptation between geographic regions using a similar sensor setup. The \emph{Day $\to$ Night} scenario evaluates adaptation to a low light scenario. Both tasks are extracted from the nuScenes dataset, which provides a significant challenge for 3D modeling due to its low point density.
The \emph{A2D2 $\to$ SemanticKITTI} (A2D2 $\to$ SK) task is a challenging dataset adaptation task, since both 2D and 3D sensor systems significantly differ in terms of resolution, field of view, and LiDAR beam structure.  The \emph{VirtualKITTI $\to$ SemanticKITTI} (VK $\to$ SK) explores a virtual-to-real adaptation, aiming to study the adaptability from generated 2D and 3D data towards real data, where the generated point clouds are randomly sampled points from depth maps of the generated scenes. The different classes are merged to six classes (ten for the A2D2-SemanticKITTI task).

\subsection{Implementation Details}
In general, we follow the setup and hyperparameters provided by xMuda \cite{jaritz2022cross}, however, we adjusted several important parameters, as detailed in the following paragraph.
The pipeline consists of a pre-trained, frozen vision foundation model with a trainable linear head, a fusion network and a 3D encoder based on a U-Net \cite{ronneberger2015unet} style 3D SparseConvNet \cite{graham2018sparseconvnet}.
We used batch size 24 for training. Further, for the training of the linear 2D classifier and the fusion network, we reduce the learning rate to $1 \times 10^{-3}$.
We set the learning rate for the 3D model to $3 \times 10^{-3}$. 
For the pseudo-label loss, we used $\lambda_{PL} = 1$ for all datasets. We set the source and target alignment coefficients $\lambda_S$ and $\lambda_T$  to 1 and 0.1 for the NuScenes tasks and 0.5 and 0.5 for the A2D2-SemanticKITTI and VirtualKITTI-SemanticKITTI task, respectively. 
The fusion modality guidance $\lambda$ is set to $1$ for adaptation in daylight target domain tasks and $0$ for night tasks to bias the fusion on modalities that are more robust in these lighting conditions. 
  
\paragraph{Vision Foundation Models.}
In our experiments we employ the AM-Radio \cite{ranzinger2024radio} VFM version 2.5-L as our primary model (patch size 16), chosen for its strong linear probing capabilities. 
To show the generalization of our approach, we also ablate with DINOv2~\cite{oquab2023dinov2}.
we observe minimal performance fluctuation. 
\paragraph{High Resolution 2D Features.}
Since ViT Transformers operate on image patches, we apply two general strategies to obtain higher resolution pixel-level features, which are essential for dense prediction tasks such as semantic segmentation. We follow \cite{liu2023seal, sautier2022slidr} and apply cropout-resize with higher resolution in order to maximize the resolution for the patch-wise feature extraction for the ViT-encoder. Second, we add bilinear interpolation following \cite{puy2024threepillars, peng2023learning} to retrieve interpolated pixel-level features.

\paragraph{Fusion Network.}
We use an MLP with two hidden layers, each followed by batch normalization, GeLU nonlinearity, and a dropout layer. We choose the hidden dimensions of the same dimension as the 2D input, i.e., 1024 for the AM-RADIO VFM. For the input we linearly project the 3D features to the 2D VFM dimension to reduce the bias from the large dimension gap between the 2D and 3D feature dimensions.

\subsection{Experimental Results}

\begin{table*}
\centering
\small
\resizebox{!}{0.23\linewidth}{\begin{tabular}{l|c|ccc|ccc|ccc|ccc|c}
\toprule
      &  & \multicolumn{3}{c}{USA $\to$ Singapore} & \multicolumn{3}{c}{Day $\to$ Night} & \multicolumn{3}{c}{VK $\to$ SK}  & \multicolumn{3}{c}{A2D2 $\to$ SK} &  \\ 
Method & VFM & 2D & 3D & 2D3D & 2D  & 3D & 2D3D &  2D & 3D & 2D3D  &  2D & 3D & 2D3D & Avg \\  \midrule \midrule
Source & & 58.4 & 62.8 & 68.2 & 47.8 & 68.8 & 63.3 & 26.8  & 42.0 & 42.2 & 34.2 & 35.9 & 40.4 & 49.2 \\ 
Target &  & 75.4 &  76.0 & 79.6 & 61.5 & 69.8 & 69.2 &66.3 & 78.4 & 80.1 & 59.3 & 71.9 & 73.6 & 71.8\\ 
\midrule
DsCML~\cite{peng2021sparsetodense}&  & 65.6& 56.2& 66.1& 50.9 &49.3& 53.2& 38.4 &38.4& 45.5& 39.6 &45.1& 44.5 & 49.4 \\ 
DsCML\textsubscript{PL}~\cite{peng2021sparsetodense} & & 65.6 & 57.5 & 66.9 & 51.4 & 49.8 & 53.8 & 39.6 & 41.8  &42.2 & 46.8  &51.8 & 52.4 & 51.6\\ 
xMUDA~\cite{jaritz2022cross} & & 64.4 & 63.2 & 69.4 & 55.5 & 69.2 & 67.4 & 42.1 & 46.7 & 48.2 & 38.3 & 46.0 & 44.0 & 54.5 \\ 
BFtD~\cite{wu2023_bidrectional_fusion}&  & 63.7 & 62.2 & 69.4 & 57.1 & 70.4 & 68.3 &  41.5 & 45.5 & 51.5 & 40.5 & 44.4 & 48.7 & 55.3\\
SSE~\cite{zhang2022exclusivelearning}&  & 64.9 & 63.9 & 69.2 & 62.8 & 69.0 & 68.9 &  45.9 & 40.0 & 49.6 & 44.5 & 46.8 & 48.4 & 56.2\\  
XMUDA\textsubscript{PL}~\cite{jaritz2022cross} & & 67.0 & 65.4 & 71.2 & 57.6 & 69.9 & 64.4 & 45.8 & 51.0 & 52.0 & 41.2 & 49.8 & 47.5 & 56.9 \\
SSE\textsubscript{PL}~\cite{zhang2022exclusivelearning}&  & 66.9 & 64.4 &70.6 & 59.1 & 67.0 & 66.3 & 47.2 & 53.5 & 55.2 & 45.9 & 51.5 & 52.5 & 58.3\\
FtD++~\cite{wu2024_fusion_then_distill2}&  &  69.7& 64.6& 69.8 & 68.8 & 69.6 & 71.0 & 51.0 & 44.0 &52.6 &48.8 &46.2 &51.1  & 58.9\\ 
BFtD\textsubscript{PL}~\cite{wu2023_bidrectional_fusion}&  &  65.9 & 66.0 & 71.3 & 60.6 & 70.0 & 66.6 & 48.6 & 55.4 & 57.5 & 42.6 & 53.7 & 52.7 & 59.2\\
MM2D3D~\cite{cardace2023exploiting} & & 71.7 & 66.8 & 72.4 & \underline{70.5} & 70.2 & \underline{72.1} & 53.4 & 50.3 & 56.5 & 42.3 & 46.1 & 46.2 & 59.9 \\ 
FtD++\textsubscript{PL}~\cite{wu2024_fusion_then_distill2} &  & 71.7 & 65.5 & 72.3 & 68.9 & 70.3 &  71.8 & 52.9 & 51.2 & 57.8 & 51.4 & 49.7 & 54.8 & 61.5\\ 
MM2D3\textsubscript{PL}~\cite{cardace2023exploiting} & & 74.3 & 68.3 &74.9 & \textbf{71.3} &69.6 &\textbf{72.2} & 55.4 & 55.0 & 59.7  &46.4 &48.7 &50.7 & 62.2 \\

LTA-SAM\textsubscript{PL}~\cite{peng2023learning}& \cmark  & - & \textbf{73.6} & - & - & \textbf{70.5} & - & - & \underline{64.9} & - & - & 52.1 & - & -\\ 
VFM-BOOST\textsubscript{PL}~\cite{xu2024visual}& \cmark  & 70.0 &65.6 &72.3 &60.6 & \textbf{70.5} &66.5 &57.2 &52.0 &61.0 &45.0& 52.3& 50.0 & 60.3\\ 
UniDSeg~\cite{wu2024unidseg}& \cmark & 67.2 &67.6 &72.9 & 63.2 & 71.2 & 71.2 & 60.5 &50.9 &62.2 &50.7& \underline{55.4}& 57.5 & 62.5 \\ 
\midrule
\textbf{Ours} & \cmark & \underline{74.4} & 67.9 & \underline{75.4} & 67.9 & 68.1 & 70.3 & \underline{70.1} & 64.5 & \underline{70.7} & \underline{60.3} & 54.9 & \underline{63.1} & \underline{67.2}\\
\textbf{Ours\textsubscript{PL}}& \cmark  & \textbf{76.1} & \underline{70.5} & \textbf{76.2} & 69.2 & 69.0 & 70.4 & \textbf{72.1} & \textbf{68.6} & \textbf{72.1} & \textbf{62.3} & \textbf{57.8} & \textbf{63.3} & \textbf{69.0}\\
\midrule
\bottomrule
\end{tabular}}
\caption{Quantitative (mIoU) results (\textbf{best}, \underline{second}). $P_L$ denotes two-stage training with pseudo-labels. For 2D we report the result of our fusion network. 2D3D denotes the softmax average of the fusion and 3D head. Source and Target: The baseline xMUDA implementation is trained either on source data or on target data, which serves as the lower and upper bound.} 
\label{tab:main_results}
\end{table*}
We evaluate our method on the four commonly used 3D UDA tasks for semantic segmentation.
Following previous works in the field~\cite{jaritz2019xmuda, cardace2023exploiting, peng2021sparsetodense}, we evaluate the performance on the test set using the checkpoint that achieved the best IoU score on the validation set.

In Table~\ref{tab:main_results} we report the results for our method and comparison with all other baselines and state-of-the-art methods. 
Generally, our method improves in most scenarios when compared with other approaches. 
We observe an impressive improvement of $6.5\%$ on average (over all tasks), when compared with the strongest baseline UniDSeg~\cite{wu2024unidseg}, which prompt-tunes a VLM. 
On individual tasks, we also find that our method generally performs well. 
When comparing with the other approaches that also employ the recently proposed VFMs, we see a positive trend and we outperform the state-of-the-art VFM-BOOST~\cite{xu2024visual} by an average of $7.5\%,$ whereas obtaining up to $14.2\%$ gains on the adaptation task of VirtualKITTI to SemanticKITTI.
Similarly, we find that our method obtains an improvement over the other VFM-guided method (LTA-SAM~\cite{peng2023learning}) of over $5\%$ and $3\%$ in the adaptation scenarios of A2D2 $\to$ SemanticKITTI and VirtualKITTI $\to$ SemanticKITTI, while remaining competitive on the other two tasks. 
These results highlight the benefits of our proposed fusion scheme and biasing of this fusion according to the downstream task of interest. 
In Table~\ref{tab:main_results} we observe that our method fares better on most tasks but remains competitive on adaptation from Day $\to$ Night scenario, where we rank second by a small margin of $0.8\%$ on average, when comparing to FtD++. 
FtD++, uses additional distillation to preserve the domain-specific attributes which helps them to perform well on the task of Day $\to$ Night adaptation, whereas our method outperforms them on all other scenarios, without requiring an additional distillation. 
Overall, we outperform FtD++ by an average of $7.5\%$. These strong results highlight the two main aspects of our method: the effective VFM utilization and the proposed modality guided fusion. 

\begin{figure*}[h!]
    \centering
    \includegraphics[width=0.98\linewidth]{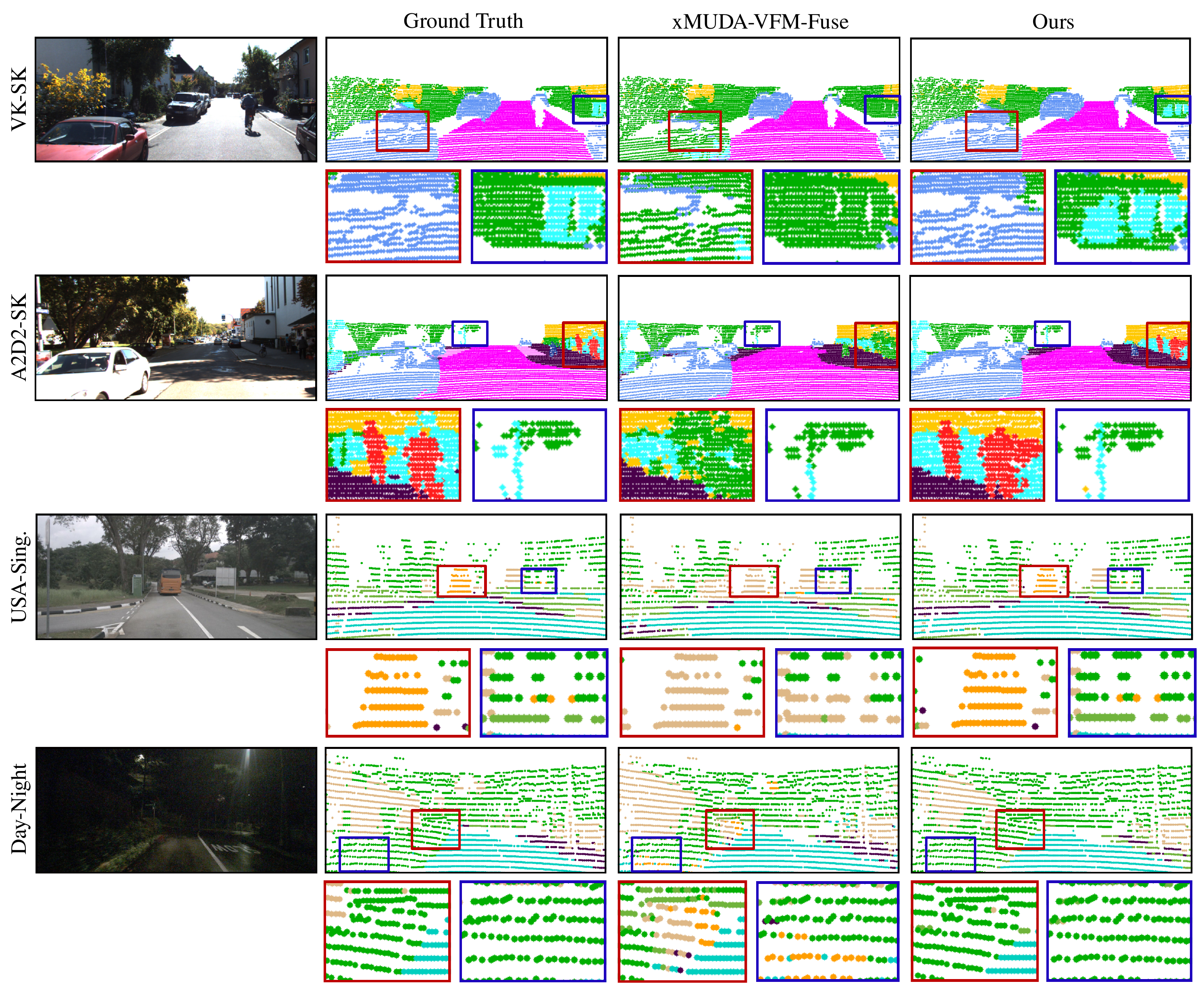}
        \caption{Qualitative comparison of our method on an example from each dataset. We show the softmax average of our fusion and 3D head. Boxes mark locations of interest with zoom-in below. Multiple traffic participants are not detected by xMUDA-VFM-Fuse; \textbf{VK~$\to$~SK}: A car is incorrectly identified as nature; \textbf{A2D2~$\to$~SK}: Two persons are missed; \textbf{USA~$\to$~Sing.} A bus is wrongly identified as a manmade structure. Our method correctly identifies these traffic participants likely due to our stronger reliance on the well-generalizing VFM features. \textbf{Day~$\to$~Night} xMUDA-VFM-Fuse detects false positive vehicles, a potential sign of overreliance on visual features in low-light conditions which can be avoided with our proposed fusion regularization.}
        \label{fig:qualitative_results}
\end{figure*}
\begin{figure}[ht]
    \centering
    \includegraphics[width=\columnwidth]{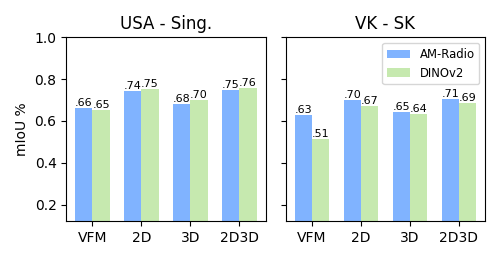}
    \caption{Comparison of current SOTA VFMs on USA $\to$ Sing. and VK $\to$ SK. 
             We report the mIoU \% for our main heads including the VFM head utilized for the fusion regularization.}
    \label{fig:vfm_comparison}
\end{figure}
\begin{table}
  \centering
\setlength{\tabcolsep}{3pt}
\begin{tabular}{l|ccc|ccc}
\toprule
      & \multicolumn{3}{c}{Day $\to$ Night} & \multicolumn{3}{c}{VK $\to$ SK}\\ 
Fusion & 2D & 3D & 2D3D &   2D & 3D & 2D3D  \\  \midrule\midrule
Vanilla &  64.8 & 68.9 & 68.3 & 68.3 & 61.1 & 68.5 \\
+ MG & 64.3 & 67.8 & 67.8 & 68.8 & 64.2 & 69.7\\
MLP & 65.9 & 67.5 &  68.8 & 69.7 & 63.2 & 69.2 \\ 
MLP + SymAl & 66.1 & 67.5 & 68.9 &  69.9 & 64.0 & 69.9 \\
\midrule
MLP + MG (ours) & 67.9 & 68.1 & 70.3 & 70.1 & 64.5 & 70.7 \\ 
\bottomrule
\bottomrule
\end{tabular}
\caption{Fusion Ablation Study. Modality-guided (MG) fusion with an MLP is compared against vanilla fusion and symmetric alignment (SymAl) from both the 2D and 3D networks. Note, that the vanilla fusion here is conducted slightly differently than the xMUDA-Fuse method since we first project the 3D features to the dimension of the 2D features.}
\label{tab:fusion_study}
\end{table}

\subsection{Ablation Study}
\label{sec:ablation_stud}
In this subsection, we provide additional experiments validating the effectiveness of our method.

\paragraph{Component ablation study.}
\label{para:fusion_study}
We ablate our fusion adaptation method starting from a lightweight fusion proposed in \cite{jaritz2019xmuda}, which consists of 2D and 3D feature concatenation followed by a projection layer and ReLU nonlinearity and evaluated the effectiveness of our proposed modality guidance.  In addition, we ablate the utilization of an MLP instead of a single layered fusion and also report the impact of the modality guidance (MG) on the MLP. For comparison, we also provide results for a symmetric alignment (SymAl) where we align the fusion from both the VFM and 3D network in a symmetric way. This means, we added another mimicry head to the fusion network such that each modality (VFM, 3D) has their respective student on the fusion network.

The results in Table~\ref{tab:fusion_study} show practically no gains when applying the guidance on the vanilla fusion, as its simple structure may prevent it from effectively responding to the modality guidance. Discernible improvements can be achieved when employing an MLP instead of a single layered fusion. The MLP fusion is improved when aligning the fusion from both the VFM and the 3D teacher, suggesting that a symmetric regularization is worthy in the absence of any priors regarding the more robust modality. However, for our evaluated tasks the full potential can be harnessed when the fusion head is guided toward the VFM on daylight target data and for the night task toward the 3D network.

\paragraph{Generalization beyond RADIO.}
\label{para:ablation study}
In Figure~\ref{fig:vfm_comparison} we evaluate our method by using DINOv2. 
We see that our method can generalize across VFMs.
With the DINOv2 backbone, we observe an improvement of $1.3\%$ for the USA $\to$ Singapore task, while we observe a degradation of $1.9\%$ on the VirtualKITTI $\to$ Semantic KITTI adaptation.
These results highlight the generalization ability of our method beyond AM-RADIO.
Further, as VFMs advance, our method can also directly benefit and obtain further performance improvements.

\paragraph{xMUDA Fusion Ablation.}
In this experiment, we compare our method with xMUDA variants including their fusion variant that we implemented with our utilized VFM. The xMUDA fusion variant has only a single classifier on top of a linear layer and a ReLu nonlinearity. As input serves the feature concatenation of the 2D and 3D features. The supervision is only applied to the fusion classifier. Further, 2D and 3D stream encoders are aligned by the fusion classifier via a mimicry head.
\label{para:fusion_baseline}
\begin{table}[ht]
  \centering
\begin{tabular}{l| c|c}
\toprule
     2D3D  & USA $\to$ Sing.  & A2D2 $\to$ SK\\ 
     \midrule \midrule
xMUDA \cite{jaritz2022cross}   &69.2 &  44.0 \\
xMUDA-Fuse \cite{jaritz2022cross} &  69.3 & 42.6  \\
xMUDA-VFM-Fuse &  73.4  & 61.4 \\
\midrule
Ours  &  75.4 & 63.1 \\ 
\bottomrule
\bottomrule
\end{tabular}
\caption{Comparison of xMUDA-Fuse variants with and without VFM backbone. Our method improves over other methods outlining the need for fusion adaptation strategies for VFM utilization. }

\label{tab:fusion_alignment_symetry}
\end{table}
The results of this ablation are presented in Table~\ref{tab:fusion_alignment_symetry}.
We find that our proposed fusion schema fares better than the fusion proposed by xMUDA. 
A notable result is obtained by comparing the xMUDA method and replacing the 2D feature extractor with the Radio VFM (employed in our work). 
We find that we outperform their method by $2\%$ and $1.7\%$ on the two adaptation scenarios we test.

\paragraph{Qualitative Results.}

\Cref{fig:qualitative_results} depicts illustrative examples of our method and a comparison to  xMUDA-VFM-Fuse. While the comparison implies similar performance for USA~$\to$ Singapore and A2D2~$\to$~SK task, our method overall produces smoother segmentation masks compared to xMUDA-VFM-Fuse, while xMUDA-VFM-Fuse more often correctly classifies occluded points (resulting from the sensor system layout) which are especially present in the SemanticKITTI dataset and can be observed on most borders of close objects, well observable in the left vehicle in the VK~$\to$~SK and the A2D2~$\to$~SK example.

\paragraph{Limitations}
The fusion guidance relies on a predefined modality preference informed by environmental conditions (e.g., prioritizing LiDAR under low-light and RGB in daylight). As such, it may be less effective in ambiguous conditions. In future work, we plan to explore mechanisms for estimating modality reliability at a per-point level, enabling flexible and fine-grained guidance based on spatial and semantic context.
\section{Conclusion}

We present a method that effectively leverages vision foundation models (VFMs) for 3D unsupervised domain adaptation (UDA) by introducing an adaptive fusion strategy informed by environmental conditions.
Specifically, we show that biasing the fusion toward the more reliable modality, based on lighting conditions, can enhance adaptation performance. 
We extensively evaluate our proposed method on four commonly employed UDA benchmarks and demonstrate strong improvements over existing state-of-the-art methods. These results offer insights into how VFMs can be effectively integrated into multi-modal learning and highlight the potential of adaptive fusion schemes. We believe that fusion modules remain a promising direction for addressing challenges such as sensor misalignment, failures, and varying environmental conditions.

{
    \small
    \bibliographystyle{ieeenat_fullname}
    \bibliography{main}

\begin{thebibliography}{53}
\providecommand{\natexlab}[1]{#1}
\providecommand{\url}[1]{\texttt{#1}}
\expandafter\ifx\csname urlstyle\endcsname\relax
  \providecommand{\doi}[1]{doi: #1}\else
  \providecommand{\doi}{doi: \begingroup \urlstyle{rm}\Url}\fi

\bibitem[Bao et~al.(2022)Bao, Dong, Piao, and Wei]{bao2021beit}
Hangbo Bao, Li Dong, Songhao Piao, and Furu Wei.
\newblock Beit: Bert pre-training of image transformers.
\newblock In \emph{ICLR}, 2022.

\bibitem[Barrera et~al.(2021)Barrera, Beltr{\'a}n, Guindel, Iglesias, and Garc{\'\i}a]{barrera2021cycle}
Alejandro Barrera, Jorge Beltr{\'a}n, Carlos Guindel, Jose~Antonio Iglesias, and Fernando Garc{\'\i}a.
\newblock Cycle and semantic consistent adversarial domain adaptation for reducing simulation-to-real domain shift in lidar bird's eye view.
\newblock In \emph{ITSC}, pages 3081--3086, 2021.

\bibitem[Behley et~al.(2019)Behley, Garbade, Milioto, Quenzel, Behnke, Stachniss, and Gall]{behley2019semantickitti}
Jens Behley, Martin Garbade, Andres Milioto, Jan Quenzel, Sven Behnke, Cyrill Stachniss, and J\"urgen Gall.
\newblock {SemantickITTI: A dataset for semantic scene understanding of LiDAR sequences}.
\newblock In \emph{CVPR}, pages 9297--9307, 2019.

\bibitem[Br{\"o}dermann et~al.(2024)Br{\"o}dermann, Sakaridis, Fu, and Van~Gool]{brodermann2024condition}
Tim Br{\"o}dermann, Christos Sakaridis, Yuqian Fu, and Luc Van~Gool.
\newblock Condition-aware multimodal fusion for robust semantic perception of driving scenes.
\newblock \emph{arXiv preprint arXiv:2410.10791}, 2024.

\bibitem[Caesar et~al.(2020)Caesar, Bankiti, Lang, Vora, Liong, Xu, Krishnan, Pan, Baldan, and Beijbom]{caesar2020nuscenes}
Holger Caesar, Varun Bankiti, Alex~H Lang, Sourabh Vora, Venice~Erin Liong, Qiang Xu, Anush Krishnan, Yu Pan, Giancarlo Baldan, and Oscar Beijbom.
\newblock nuscenes: A multimodal dataset for autonomous driving.
\newblock In \emph{CVPR}, pages 11621--11631, 2020.

\bibitem[Cao et~al.(2024)Cao, Xu, Yang, Yin, Yuan, and Xie]{cao2024mopa}
Haozhi Cao, Yuecong Xu, Jianfei Yang, Pengyu Yin, Shenghai Yuan, and Lihua Xie.
\newblock Mopa: Multi-modal prior aided domain adaptation for 3d semantic segmentation.
\newblock In \emph{ICRA}, pages 9463--9470, 2024.

\bibitem[Cardace et~al.(2023)Cardace, Ramirez, Salti, and Di~Stefano]{cardace2023exploiting}
Adriano Cardace, Pierluigi~Zama Ramirez, Samuele Salti, and Luigi Di~Stefano.
\newblock Exploiting the complementarity of 2d and 3d networks to address domain-shift in 3d semantic segmentation.
\newblock In \emph{CVPR Workshop}, pages 98--109, 2023.

\bibitem[Caron et~al.(2021)Caron, Touvron, Misra, J{\'e}gou, Mairal, Bojanowski, and Joulin]{caron2021emerging_dinov1}
Mathilde Caron, Hugo Touvron, Ishan Misra, Herv{\'e} J{\'e}gou, Julien Mairal, Piotr Bojanowski, and Armand Joulin.
\newblock Emerging properties in self-supervised vision transformers.
\newblock In \emph{CVPR}, pages 9650--9660, 2021.

\bibitem[Chen et~al.(2020)Chen, Wei, Kumar, and Ma]{chen2020self}
Yining Chen, Colin Wei, Ananya Kumar, and Tengyu Ma.
\newblock Self-training avoids using spurious features under domain shift.
\newblock In \emph{NeurIPS}, pages 21061--21071, 2020.

\bibitem[Choi et~al.(2018)Choi, Choi, Kim, Ha, Kim, and Choo]{choi2018stargan}
Yunjey Choi, Minje Choi, Munyoung Kim, Jung-Woo Ha, Sunghun Kim, and Jaegul Choo.
\newblock Stargan: Unified generative adversarial networks for multi-domain image-to-image translation.
\newblock In \emph{CVPR}, pages 8789--8797, 2018.

\bibitem[Dosovitskiy(2021)]{dosovitskiy2021image}
Alexey Dosovitskiy.
\newblock An image is worth 16x16 words: Transformers for image recognition at scale.
\newblock In \emph{ICLR}, 2021.

\bibitem[Gaidon et~al.(2016)Gaidon, Wang, Cabon, and Vig]{gaidon2016virtualkitti}
Adrien Gaidon, Qiao Wang, Yohann Cabon, and Eleonora Vig.
\newblock Virtual worlds as proxy for multi-object tracking analysis.
\newblock In \emph{CVPR}, pages 4340--4349, 2016.

\bibitem[Ganin et~al.(2016)Ganin, Ustinova, Ajakan, Germain, Larochelle, Laviolette, March, and Lempitsky]{ganin2016domain}
Yaroslav Ganin, Evgeniya Ustinova, Hana Ajakan, Pascal Germain, Hugo Larochelle, Fran{\c{c}}ois Laviolette, Mario March, and Victor Lempitsky.
\newblock Domain-adversarial training of neural networks.
\newblock \emph{{Journal of Machine Learning Research}}, 17\penalty0 (59):\penalty0 1--35, 2016.

\bibitem[Geyer et~al.(2020)Geyer, Kassahun, Mahmudi, Ricou, Durgesh, Chung, Hauswald, Pham, M{\"u}hlegg, Dorn, et~al.]{geyer2020a2d2}
Jakob Geyer, Yohannes Kassahun, Mentar Mahmudi, Xavier Ricou, Rupesh Durgesh, Andrew~S Chung, Lorenz Hauswald, Viet~Hoang Pham, Maximilian M{\"u}hlegg, Sebastian Dorn, et~al.
\newblock A2d2: Audi autonomous driving dataset.
\newblock \emph{arXiv preprint arXiv:2004.06320}, 2020.

\bibitem[Graham et~al.(2018)Graham, Engelcke, and Van Der~Maaten]{graham2018sparseconvnet}
Benjamin Graham, Martin Engelcke, and Laurens Van Der~Maaten.
\newblock 3d semantic segmentation with submanifold sparse convolutional networks.
\newblock In \emph{CVPR}, pages 9224--9232, 2018.

\bibitem[He et~al.(2016)He, Zhang, Ren, and Sun]{resnet}
Kaiming He, Xiangyu Zhang, Shaoqing Ren, and Jian Sun.
\newblock Deep residual learning for image recognition.
\newblock In \emph{CVPR}, pages 770--778, 2016.

\bibitem[Hoffman et~al.(2018)Hoffman, Tzeng, Park, Zhu, Isola, Saenko, Efros, and Darrell]{hoffman2018cycada}
Judy Hoffman, Eric Tzeng, Taesung Park, Jun-Yan Zhu, Phillip Isola, Kate Saenko, Alexei Efros, and Trevor Darrell.
\newblock Cycada: Cycle-consistent adversarial domain adaptation.
\newblock In \emph{ICML}, pages 1989--1998, 2018.

\bibitem[Jaritz et~al.(2020)Jaritz, Vu, de~Charette, Wirbel, and P{\'e}rez]{jaritz2019xmuda}
Maximilian Jaritz, Tuan-Hung Vu, Raoul de Charette, Emilie Wirbel, and Patrick P{\'e}rez.
\newblock {xMUDA}: Cross-modal unsupervised domain adaptation for {3D} semantic segmentation.
\newblock In \emph{CVPR}, 2020.

\bibitem[Jaritz et~al.(2022)Jaritz, Vu, de~Charette, Wirbel, and P{\'e}rez]{jaritz2022cross}
Maximilian Jaritz, Tuan-Hung Vu, Raoul de Charette, Emilie Wirbel, and Patrick P{\'e}rez.
\newblock Cross-modal learning for domain adaptation in {3D} semantic segmentation.
\newblock In \emph{PAMI}, 2022.

\bibitem[Jia et~al.(2021)Jia, Yang, Xia, Chen, Parekh, Pham, Le, Sung, Li, and Duerig]{jia2021scaling}
Chao Jia, Yinfei Yang, Ye Xia, Yi-Ting Chen, Zarana Parekh, Hieu Pham, Quoc Le, Yun-Hsuan Sung, Zhen Li, and Tom Duerig.
\newblock Scaling up visual and vision-language representation learning with noisy text supervision.
\newblock In \emph{ICML}, pages 4904--4916, 2021.

\bibitem[Kirillov et~al.(2023)Kirillov, Mintun, Ravi, Mao, Rolland, Gustafson, Xiao, Whitehead, Berg, Lo, et~al.]{kirillov2023segment}
Alexander Kirillov, Eric Mintun, Nikhila Ravi, Hanzi Mao, Chloe Rolland, Laura Gustafson, Tete Xiao, Spencer Whitehead, Alexander~C Berg, Wan-Yen Lo, et~al.
\newblock Segment anything.
\newblock In \emph{CVPR}, pages 4015--4026, 2023.

\bibitem[Laine and Aila(2017)]{laine2016temporal}
Samuli Laine and Timo Aila.
\newblock Temporal ensembling for semi-supervised learning.
\newblock \emph{ICLR}, 2017.

\bibitem[Lee et~al.(2013)]{lee2013pseudo}
Dong-Hyun Lee et~al.
\newblock Pseudo-label: The simple and efficient semi-supervised learning method for deep neural networks.
\newblock In \emph{ICML Workshop}, page 896, 2013.

\bibitem[Li et~al.(2023)Li, Dai, Han, and Ding]{li2023mseg3d}
Jiale Li, Hang Dai, Hao Han, and Yong Ding.
\newblock Mseg3d: Multi-modal 3d semantic segmentation for autonomous driving.
\newblock In \emph{CVPR}, pages 21694--21704, 2023.

\bibitem[Li et~al.(2018)Li, Wang, Shi, Hou, and Liu]{li2018adaptive}
Yanghao Li, Naiyan Wang, Jianping Shi, Xiaodi Hou, and Jiaying Liu.
\newblock Adaptive batch normalization for practical domain adaptation.
\newblock \emph{Pattern Recognition}, 80:\penalty0 109--117, 2018.

\bibitem[Liu et~al.(2021{\natexlab{a}})Liu, Wang, and Long]{liu2021cycle}
Hong Liu, Jianmin Wang, and Mingsheng Long.
\newblock Cycle self-training for domain adaptation.
\newblock \emph{NeurIPS}, pages 22968--22981, 2021{\natexlab{a}}.

\bibitem[Liu et~al.(2021{\natexlab{b}})Liu, Luo, Cai, Yu, Ke, Junior, Gon{\c{c}}alves, and Li]{liu2021auda}
Wei Liu, Zhiming Luo, Yuanzheng Cai, Ying Yu, Yang Ke, Jos{\'e}~Marcato Junior, Wesley~Nunes Gon{\c{c}}alves, and Jonathan Li.
\newblock Adversarial unsupervised domain adaptation for 3d semantic segmentation with multi-modal learning.
\newblock \emph{ISPRS Journal of Photogrammetry and Remote Sensing}, 176:\penalty0 211--221, 2021{\natexlab{b}}.

\bibitem[Liu et~al.(2023)Liu, Kong, Cen, Chen, Zhang, Pan, Chen, and Liu]{liu2023seal}
Youquan Liu, Lingdong Kong, Jun Cen, Runnan Chen, Wenwei Zhang, Liang Pan, Kai Chen, and Ziwei Liu.
\newblock Segment any point cloud sequences by distilling vision foundation models.
\newblock In \emph{NeurIPS}, pages 37193--37229, 2023.

\bibitem[Liu et~al.(2022)Liu, Mao, Wu, Feichtenhofer, Darrell, and Xie]{convnet}
Zhuang Liu, Hanzi Mao, Chao-Yuan Wu, Christoph Feichtenhofer, Trevor Darrell, and Saining Xie.
\newblock A convnet for the 2020s.
\newblock In \emph{CVPR}, pages 11976--11986, 2022.

\bibitem[Mei et~al.(2020)Mei, Zhu, Zou, and Zhang]{mei2020instance}
Ke Mei, Chuang Zhu, Jiaqi Zou, and Shanghang Zhang.
\newblock Instance adaptive self-training for unsupervised domain adaptation.
\newblock In \emph{ECCV}, pages 415--430, 2020.

\bibitem[Michele et~al.(2024)Michele, Boulch, Puy, Vu, Marlet, and Courty]{michele2024saluda}
Bj{\"o}rn Michele, Alexandre Boulch, Gilles Puy, Tuan-Hung Vu, Renaud Marlet, and Nicolas Courty.
\newblock Saluda: Surface-based automotive lidar unsupervised domain adaptation.
\newblock In \emph{3DV}, pages 421--431, 2024.

\bibitem[Mirza et~al.(2022)Mirza, Micorek, Possegger, and Bischof]{mirza2022norm}
M~Jehanzeb Mirza, Jakub Micorek, Horst Possegger, and Horst Bischof.
\newblock The norm must go on: Dynamic unsupervised domain adaptation by normalization.
\newblock In \emph{CVPR}, pages 14765--14775, 2022.

\bibitem[Morerio et~al.(2018)Morerio, Cavazza, and Murino]{morerio2018minimal}
Pietro Morerio, Jacopo Cavazza, and Vittorio Murino.
\newblock Minimal-entropy correlation alignment for unsupervised deep domain adaptation.
\newblock In \emph{ICLR}, 2018.

\bibitem[Oquab et~al.(2023)Oquab, Darcet, Moutakanni, Vo, Szafraniec, Khalidov, Fernandez, Haziza, Massa, El-Nouby, et~al.]{oquab2023dinov2}
Maxime Oquab, Timoth{\'e}e Darcet, Th{\'e}o Moutakanni, Huy Vo, Marc Szafraniec, Vasil Khalidov, Pierre Fernandez, Daniel Haziza, Francisco Massa, Alaaeldin El-Nouby, et~al.
\newblock Dinov2: Learning robust visual features without supervision.
\newblock \emph{arXiv preprint arXiv:2304.07193}, 2023.

\bibitem[Peng et~al.(2021)Peng, Lei, Li, Zhang, and Guo]{peng2021sparsetodense}
Duo Peng, Yinjie Lei, Wen Li, Pingping Zhang, and Yulan Guo.
\newblock Sparse-to-dense feature matching: Intra and inter domain cross-modal learning in domain adaptation for 3d semantic segmentation.
\newblock In \emph{CVPR}, pages 7108--7117, 2021.

\bibitem[Peng et~al.(2024)Peng, Chen, Qiao, Kong, Liu, Wang, Zhu, and Ma]{peng2023learning}
Xidong Peng, Runnan Chen, Feng Qiao, Lingdong Kong, Youquan Liu, T Wang, X Zhu, and Y Ma.
\newblock Learning to adapt sam for segmenting cross-domain point clouds.
\newblock In \emph{ECCV}, 2024.

\bibitem[Puy et~al.(2024)Puy, Gidaris, Boulch, Sim{\'e}oni, Sautier, P{\'e}rez, Bursuc, and Marlet]{puy2024threepillars}
Gilles Puy, Spyros Gidaris, Alexandre Boulch, Oriane Sim{\'e}oni, Corentin Sautier, Patrick P{\'e}rez, Andrei Bursuc, and Renaud Marlet.
\newblock Three pillars improving vision foundation model distillation for lidar.
\newblock In \emph{CVPR}, pages 21519--21529, 2024.

\bibitem[Radford et~al.(2021)Radford, Kim, Hallacy, Ramesh, Goh, Agarwal, Sastry, Askell, Mishkin, Clark, et~al.]{radford2021learning}
Alec Radford, Jong~Wook Kim, Chris Hallacy, Aditya Ramesh, Gabriel Goh, Sandhini Agarwal, Girish Sastry, Amanda Askell, Pamela Mishkin, Jack Clark, et~al.
\newblock Learning transferable visual models from natural language supervision.
\newblock In \emph{ICML}, pages 8748--8763, 2021.

\bibitem[Ranzinger et~al.(2024)Ranzinger, Heinrich, Kautz, and Molchanov]{ranzinger2024radio}
Mike Ranzinger, Greg Heinrich, Jan Kautz, and Pavlo Molchanov.
\newblock Am-radio: Agglomerative vision foundation model reduce all domains into one.
\newblock In \emph{CVPR}, pages 12490--12500, 2024.

\bibitem[Ronneberger et~al.(2015)Ronneberger, Fischer, and Brox]{ronneberger2015unet}
Olaf Ronneberger, Philipp Fischer, and Thomas Brox.
\newblock U-net: Convolutional networks for biomedical image segmentation.
\newblock In \emph{MICCAI}, pages 234--241, 2015.

\bibitem[Sautier et~al.(2022)Sautier, Puy, Gidaris, Boulch, Bursuc, and Marlet]{sautier2022slidr}
Corentin Sautier, Gilles Puy, Spyros Gidaris, Alexandre Boulch, Andrei Bursuc, and Renaud Marlet.
\newblock Image-to-lidar self-supervised distillation for autonomous driving data.
\newblock In \emph{CVPR}, pages 9891--9901, 2022.

\bibitem[Sun and Saenko(2016)]{sun2016deep}
Baochen Sun and Kate Saenko.
\newblock Deep coral: Correlation alignment for deep domain adaptation.
\newblock In \emph{ECCV Workshop}, pages 443--450, 2016.

\bibitem[Wu et~al.(2023)Wu, Xing, Zhang, Xie, Fan, Shi, and Qu]{wu2023_bidrectional_fusion}
Yao Wu, Mingwei Xing, Yachao Zhang, Yuan Xie, Jianping Fan, Zhongchao Shi, and Yanyun Qu.
\newblock Cross-modal unsupervised domain adaptation for 3d semantic segmentation via bidirectional fusion-then-distillation.
\newblock In \emph{ACMMM}, pages 490--498, 2023.

\bibitem[Wu et~al.(2024{\natexlab{a}})Wu, Xing, Zhang, Luo, Xie, and Qu]{wu2024unidseg}
Yao Wu, Mingwei Xing, Yachao Zhang, Xiaotong Luo, Yuan Xie, and Yanyun Qu.
\newblock Unidseg: Unified cross-domain 3d semantic segmentation via visual foundation models prior.
\newblock In \emph{NeurIPS}, pages 101223--101249, 2024{\natexlab{a}}.

\bibitem[Wu et~al.(2024{\natexlab{b}})Wu, Xing, Zhang, Xie, and Qu]{wu2024_fusion_then_distill2}
Yao Wu, Mingwei Xing, Yachao Zhang, Yuan Xie, and Yanyun Qu.
\newblock Fusion-then-distillation: Toward cross-modal positive distillation for domain adaptive 3d semantic segmentation, 2024{\natexlab{b}}.
\newblock arXiv preprint.

\bibitem[Xing et~al.(2023)Xing, Ying, Wang, Yang, and Chen]{xing2023cross}
Bowei Xing, Xianghua Ying, Ruibin Wang, Jinfa Yang, and Taiyan Chen.
\newblock Cross-modal contrastive learning for domain adaptation in 3d semantic segmentation.
\newblock In \emph{AAAI}, pages 2974--2982, 2023.

\bibitem[Xu et~al.(2024)Xu, Yang, Kong, Liu, Zhang, Zhou, and Fei]{xu2024visual}
Jingyi Xu, Weidong Yang, Lingdong Kong, Youquan Liu, Rui Zhang, Qingyuan Zhou, and Ben Fei.
\newblock Visual foundation models boost cross-modal unsupervised domain adaptation for 3d semantic segmentation, 2024.
\newblock arXiv preprint.

\bibitem[Yi et~al.(2021)Yi, Gong, and Funkhouser]{yi2021complete}
Li Yi, Boqing Gong, and Thomas Funkhouser.
\newblock Complete \& label: A domain adaptation approach to semantic segmentation of lidar point clouds.
\newblock In \emph{CVPR}, pages 15363--15373, 2021.

\bibitem[Yuan et~al.(2023)Yuan, Cheng, Zeng, Su, Liu, Yu, and Wang]{Yuan2023PrototypeGuidedMA}
Zhimin Yuan, Ming Cheng, Wankang Zeng, Yanfei Su, Weiquan Liu, Shangshu Yu, and Cheng Wang.
\newblock Prototype-guided multitask adversarial network for cross-domain lidar point clouds semantic segmentation.
\newblock \emph{IEEE Transactions on Geoscience and Remote Sensing}, pages 1--13, 2023.

\bibitem[Zagoruyko and Komodakis(2017)]{wide-resnets}
Sergey Zagoruyko and Nikos Komodakis.
\newblock Wide residual networks, 2017.

\bibitem[Zhang et~al.(2022)Zhang, Li, Xie, Li, Wang, Zhang, and Qu]{zhang2022exclusivelearning}
Yachao Zhang, Miaoyu Li, Yuan Xie, Cuihua Li, Cong Wang, Zhizhong Zhang, and Yanyun Qu.
\newblock Self-supervised exclusive learning for 3d segmentation with cross-modal unsupervised domain adaptation.
\newblock In \emph{ACMMM}, pages 3338--3346, 2022.

\bibitem[Zou et~al.(2024)Zou, Yang, Zhang, Li, Li, Wang, Wang, Gao, and Lee]{zou2024segment}
Xueyan Zou, Jianwei Yang, Hao Zhang, Feng Li, Linjie Li, Jianfeng Wang, Lijuan Wang, Jianfeng Gao, and Yong~Jae Lee.
\newblock Segment everything everywhere all at once.
\newblock In \emph{NeurIPS}, 2024.

\bibitem[Zou et~al.(2019)Zou, Yu, Liu, Kumar, and Wang]{zou2019confidence}
Yang Zou, Zhiding Yu, Xiaofeng Liu, BVK Kumar, and Jinsong Wang.
\newblock Confidence regularized self-training.
\newblock In \emph{CVPR}, pages 5982--5991, 2019.

\end{thebibliography}
}


\end{document}